\documentclass[10pt,twocolumn,letterpaper]{article}

\usepackage{cvpr}              %

\usepackage{times}
\usepackage{epsfig}
\usepackage{graphicx}
\usepackage{amsmath}
\usepackage{amssymb}
\usepackage{amsfonts}
\usepackage{amsthm}
\usepackage{url}
\usepackage{color}
\usepackage{bbm}
\usepackage{tabularx}
\usepackage{xspace}
\usepackage{caption}
\usepackage{booktabs}
\usepackage{microtype}
\usepackage{adjustbox}
\usepackage[accsupp]{axessibility}
\usepackage[outline]{contour}
\makeatletter
\@namedef{ver@everyshi.sty}{}
\makeatother
\usepackage{tikz}
\usepackage{bbm}
\usepackage{mathtools}
\newsavebox{\measurebox}

\usepackage[pagebackref,breaklinks,colorlinks]{hyperref}

\usepackage{algorithm}
\usepackage{algpseudocode}

\newcommand{\viewdir}{{\boldsymbol{\omega}_o}}
\newcommand{\lightdir}{{\boldsymbol{\omega}_i}}
\newcommand{\location}{{\mathbf{x}}}
\newcommand{\normaldir}{{\mathbf{n}}}

\newcommand{\expnumber}[2]{{#1}\mathrm{e}{#2}}

\newenvironment{packed_itemize}{
\begin{list}{\labelitemi}{\leftmargin=2em}
\vspace{-6pt}
 \setlength{\itemsep}{0pt}
 \setlength{\parskip}{0pt}
 \setlength{\parsep}{0pt}
}{\end{list}}

\usepackage[capitalize]{cleveref}
\crefname{section}{Sec.}{Secs.}
\Crefname{section}{Section}{Sections}
\Crefname{table}{Table}{Tables}
\crefname{table}{Tab.}{Tabs.}

\def\cvprPaperID{4298} %
\def\confName{CVPR}
\def\confYear{2022}

\begin{document}

\title{IRON: Inverse Rendering by Optimizing Neural SDFs and Materials \\ from Photometric Images}
\author{
Kai Zhang\textsuperscript{1}
\and Fujun Luan\textsuperscript{2}
\and Zhengqi Li\textsuperscript{3}
\and Noah Snavely\textsuperscript{1}
\smallskip
\and
\textsuperscript{1}Cornell University\qquad 
\textsuperscript{2}Adobe Research\qquad \textsuperscript{3}Google Research 
}

\maketitle
\newcommand{\kz}[1]{\noindent {\color{red} {\bf Kai says:}     {#1}}}
 \newcommand{\fl}[1]{\noindent {\color{cyan}   {\bf Fujun says:}   {#1}}}
\newcommand{\zl}[1]{\noindent {\color{blue}   {\bf Zhengqi says:} {#1}}}
\newcommand{\noah}[1]{\noindent {\color{purple} [{\bf Noah says:}   {#1}]}}

\newlength{\resLen}

\begin{abstract}

    We propose a neural inverse rendering pipeline called IRON that operates on photometric images and outputs high-quality 3D content in the format of triangle meshes and material textures readily deployable in existing graphics pipelines. Our method adopts neural representations for geometry as signed distance fields (SDFs) and materials during optimization to enjoy their flexibility and compactness, and features a hybrid optimization scheme for neural SDFs: first, optimize using a volumetric radiance field approach to
    recover correct topology, then optimize further using edge-aware physics-based surface rendering for geometry refinement and disentanglement of materials and lighting. In the second stage, we also draw inspiration from mesh-based differentiable rendering, and design a novel edge sampling algorithm for neural SDFs to further improve performance. We show that our IRON achieves significantly better inverse rendering quality compared to prior works. \footnote{Our project page is at: \url{https://kai-46.github.io/IRON-website/}.}

\end{abstract}

\begin{figure}[htb]
    \centering
    \subfloat[\centering Reconstruction of real-world objects, rendered under global illumination.]{{\includegraphics[width=0.99\columnwidth]{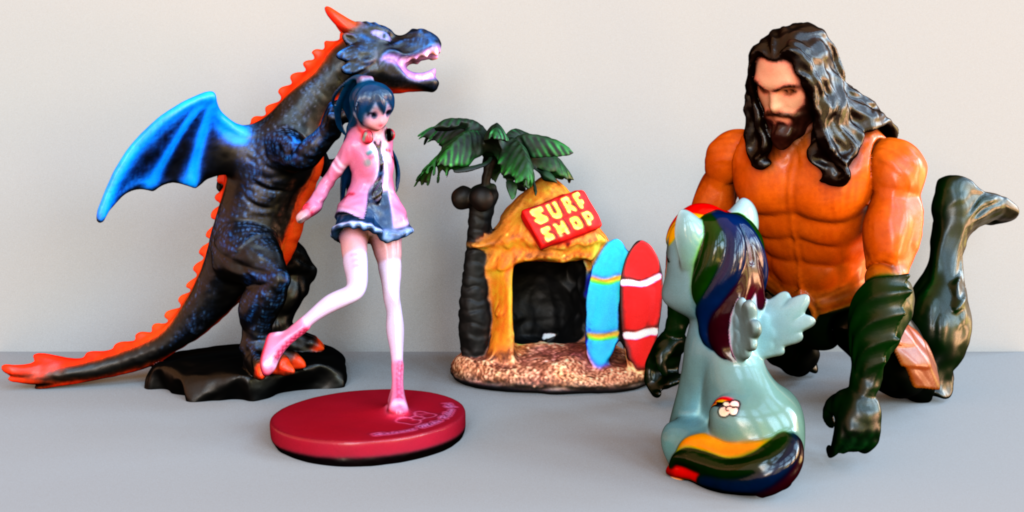} }}%
    \qquad
    \subfloat[\centering Scene editing by modifying illumination and materials, and inserting objects.]{{\includegraphics[width=0.99\columnwidth]{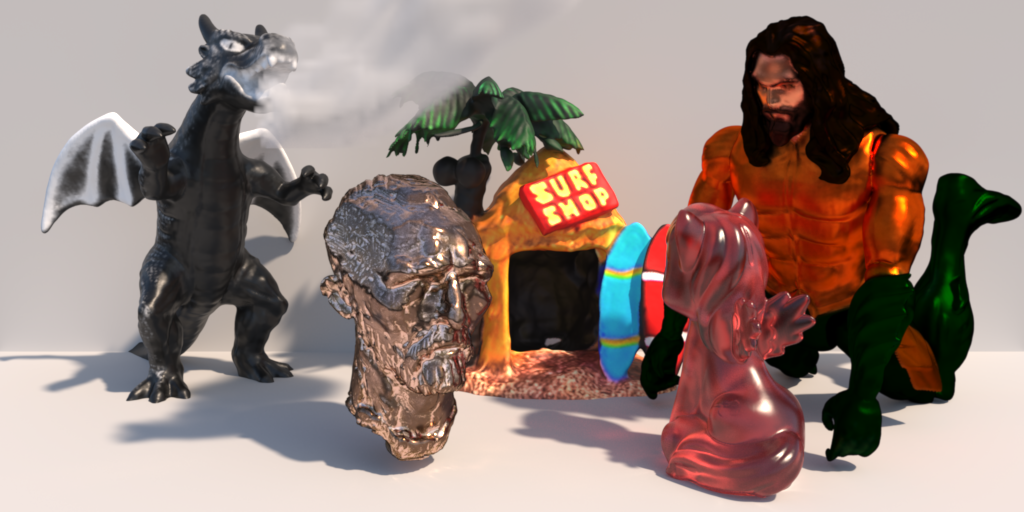} }}%
    \caption{Real-world objects reconstructed by our IRON pipeline. We show (a) re-renderings under novel global illumination via Mitsuba path tracing~\cite{Mitsuba}, and (b) scene edits that 
    include changing lighting, material BRDFs, and the scale/orientation of existing objects, and inserting new virtual objects or participating media such as the metal Van Gogh head and the heterogeneous smoke. }
    \vspace{-0.5\baselineskip}
    \label{fig:teaser}
\end{figure}

\section{Introduction}
Inverse rendering---the reconstruction of shape and appearance of real-world objects from a set of 2D input images---can enable
accessible, high-quality digitization of our world. 
One way to formulate this problem is as the inversion of rendering algorithms used in computer graphics. 
Recent advances in graphics have led to  
fully differentiable Monte Carlo path tracing methods for jointly optimizing geometry and 
BRDFs represented as triangle meshes and material textures. However, meshes 
can be difficult to optimize,
because it is non-trivial to modify their topology or maintain regularity 
during optimization. 
On the other hand, recently developed neural representations for shape~\cite{park2019deepsdf,mescheder2019occupancy} and radiance fields~\cite{mildenhall2020nerf,oechsle2019texture} demonstrate impressive success in view synthesis~\cite{mildenhall2020nerf} and shape reconstruction~\cite{yariv2020multiview,oechsle2021unisurf,wang2021neus,yariv2021volume,niemeyer2020differentiable} tasks. 
But these representations entangle material and lighting, and cannot be directly used for applications like relighting or material editing. Recent methods for decoupling neural radiance fields 
are either limited to simple shapes~\cite{zhang2021physg, boss2021nerd}, or have neural components~\cite{zhang2021nerfactor,bi2020deepreflectance} that are incompatible with standard 3D rendering and editing tools.

We address the inverse rendering problem with the objective of embracing both the flexibility and compactness of neural representations, and the convenience of meshes with material textures in downstream applications. 
We present a pipeline we call IRON---Inverse Rendering by Optimizing Neural scene components, including neural signed distance fields (SDF) and neural materials. 
IRON performs an inverse rendering optimization starting from multi-view images captured by co-locating a flashlight with a moving camera~\cite{luan2021unified,bi2020deep,bi2020deepreflectance,nam2018practical} (called \emph{photometric images} in recent work), while allowing for the export of the optimized 3D content in mesh and material texture formats readily usable by traditional graphics renderers and AR/VR applications, as shown in Fig.~\ref{fig:teaser}. At the core of IRON are four compact neural networks representing the neural SDF and materials, which we optimize with a hybrid optimization scheme: first, a volumetric radiance field optimization to recover object topology, followed by  physics-based surface rendering to refine geometric details and disentangle materials and lighting. We demonstrate superior inverse rendering quality over baseline methods targeting photometric images.

In addition, recent differentiable rendering methods in the graphics community emphasize the importance of carefully considering occlusion boundaries when performing inverse rendering with meshes, e.g., with special \emph{edge sampling} algorithms that compute accurate gradients at such edges~\cite{li2018differentiable,nimier2019mitsuba,bangaru2020warpedsampling,Zhang:2021:PSDR,cole2021differentiable,Laine2020diffrast}. In addition to our system as a whole, we propose an edge sampling algorithm defined for neural SDF representations, rather than meshes as in prior work. We show that our new edge sampling algorithm significantly improves reconstruction quality around edges.

\medskip
\noindent\textbf{Contributions.} To summarize, our key contributions are: 
\begin{packed_itemize}
    \item IRON, a neural inverse rendering pipeline for high-quality reconstruction of object shape and spatially varying materials, outperforming existing methods for photometric images.
    \item A hybrid optimization scheme for neural SDFs and materials that first optimizes a volumetric radiance field then performs edge-aware physics-based surface rendering for improved performance and compatibility with meshes and material textures. 
    \item An edge-aware rendering optimization featuring a novel edge sampling algorithm that generates unbiased gradient estimates for better optimizing neural SDFs.
\end{packed_itemize}

\section{Related work}

\noindent\textbf{Neural reconstruction and view synthesis.} Neural shape \cite{park2019deepsdf, mescheder2019occupancy} and appearance \cite{oechsle2019texture} representations based on Multi-layer Perceptrons (MLP) have been of recent interest, due to their compactness and representation power. 
Many works apply these representations to view synthesis and/or 3D reconstruction from multiple images, including NeRF~\cite{mildenhall2020nerf}, DVR~\cite{niemeyer2020differentiable}, and IDR~\cite{yariv2020multiview}. While NeRF yields view synthesis results of remarkable quality on complex scenes, its volume rendering nature 
leads to reduced quality of estimated surface geometry. In contrast, DVR and IDR, which utilize surface rendering, can reconstruct high-quality surface geometry, but are limited to relatively simple scenes compared to NeRF. Hence, UNISURF~\cite{oechsle2021unisurf}, VolSDF~\cite{yariv2021volume}, and  NeuS~\cite{wang2021neus} sought to combine the benefits of both volume- and surface-based rendering by considering surfaces as defining volumes near the surface.
Our work also utilizes neural shape and appearance, but rather than encoding appearance as a surface light field~\cite{wood2000surface} that entangles lighting and materials, we adopt a physics-based surface shading model within an inverse rendering framework. Our pipeline can output meshes and materials readily importable by existing graphics pipelines, e.g., Blender~\cite{blender}, %
for fast raytracing, object insertion, relighting, and material editing.

\medskip
\noindent\textbf{Mesh-based differentiable rendering.} Meshes are 
widely used 
in 
graphics pipelines and game engines.
There has been a surge of recent interest in fully-differentiable forward rendering methods~\cite{li2018differentiable, nimier2019mitsuba, loubet2019reparameterizing, radiative, bangaru2020warpedsampling, Zhang:2021:PSDR,cole2021differentiable,Laine2020diffrast}, 
which enable 
joint optimization of shape, material, and camera parameters from images~\cite{luan2021unified}. One 
challenge
is proper 
derivative computations at depth discontinuities
(called \emph{edge derivatives})
that arise when rendering a mesh to an image.
In addition, meshes are not 
friendly for shape optimization, due to difficulties posed in changing their topology and avoiding self-intersections~\cite{Nicolet2021Large}.
In contrast, signed distance field (SDF) representations commonly used in neural geometry methods~\cite{park2019deepsdf} parametrize the surface as the zero level set of a continuous SDF (instantiated in neural methods by an MLP), alleviating these challenges.
However, we observe that, like meshes, optimizing a neural SDF via differentiable surface rendering also requires computing both interior derivatives and edge derivatives with respect to neural network weights. IDR~\cite{yariv2020multiview} only uses interior derivatives, leading to poor performance around edges.
We introduce a method to estimate edge derivatives for better optimization of neural SDFs.

\medskip
\noindent\textbf{Inverse rendering from multiple images.} Given multiple images of a scene, inverse rendering methods seek to recover the shape, material and lighting that best explain the observed images. %
Prior work considers several capture scenarios, including:
1) static scene, static environmental lighting, moving camera~\cite{zhang2021physg,zhang2021nerfactor,boss2021nerd}, 
2) rotating scene, static environmental lighting, fixed camera~\cite{dong2014appearance,xia2016recovering}, and 
3) static scene, flashlight illumination co-located with moving camera~\cite{nam2018practical, bi2020deepreflectance, luan2021unified, bi2020deep} (in some works called \emph{photometric images}). 
Our work addresses the third case of flashlight photography because it simplifies the physics-based rendering module, while enabling high-quality inverse rendering results. Using neural SDFs and materials, we show improved results compared to baselines, while also creating easily deployed mesh and texture outputs.

\section{Method}

\smallskip
\noindent\textbf{Assumptions.} We focus on opaque objects, and consider transparent and translucent objects outside the scope of our work. We also assume that the input photometric images are captured using collocated flashlight illumination without ambient light. Finally, for 
efficiency, we ignore shadows (which are minimal in practice due to the collocated flashlight) and global illumination effects like interreflections. 

Under these assumptions, and provided with an input set of photometric images, our IRON system optimizes for neural shape and material representations consisting of four compact MLPs with positional encoding~\cite{mildenhall2020nerf, tancik2020fourier}:

\smallskip
\noindent\textbf{Neural SDF} $S_{\boldsymbol{\Theta}_s}: \boldsymbol{x} \xrightarrow{}(S, \boldsymbol{f})$ 
    maps a 3D location $\boldsymbol{x}$ to an SDF value $S$ and a 256D local geometric feature descriptor $\boldsymbol{f}$, as in recent works~\cite{oechsle2021unisurf,wang2021neus,yariv2020multiview}; this feature descriptor can then be fed into our neural material networks.

\smallskip
\noindent\textbf{Neural diffuse albedo} $\beta_{\boldsymbol{\Theta}_\beta}: (\boldsymbol{x},\boldsymbol{n},\boldsymbol{n},\boldsymbol{f}) \xrightarrow{}\beta$  outputs the diffuse albedo $\beta$ at location $\boldsymbol{x}$ given surface normal $\boldsymbol{n}$ and feature descriptor $\boldsymbol{f}$. This function plays a dual role in our optimization. In our first optimization phase, we treat $\beta$ as a neural radiance field, and include view direction as an additional parameter (in place of the second $\boldsymbol{n}$). In the second phase, we 
constrain $\beta$ to represent diffuse albedo in our edge-aware physics-based surface rendering. Our neural diffuse albedo MLP utilizes the surface normal and local geometric feature descriptors as in prior work~\cite{yariv2020multiview,oechsle2021unisurf,wang2021neus}.

\smallskip
\noindent\textbf{Neural specular albedo} $\kappa_{\boldsymbol{\Theta}_\kappa}: (\boldsymbol{x},\boldsymbol{n},\boldsymbol{f})\xrightarrow{}\kappa$ encodes spatially-varying specular albedo $\kappa$. 

\smallskip
\noindent\textbf{Neural roughness} $\alpha_{\boldsymbol{\Theta}_\alpha}: (\boldsymbol{x},\boldsymbol{n},\boldsymbol{f})\xrightarrow{}\alpha$ encodes the spatially-varying specular roughness. Small values indicate shiny surfaces, and large values less shiny.

We optimize the MLP weights $\boldsymbol{\Theta}_s,\boldsymbol{\Theta}_\beta,\boldsymbol{\Theta}_\kappa$,  and $\boldsymbol{\Theta}_\alpha$ and a scalar light intensity $L$ in a two-stage 
optimization scheme.
In the first stage, we optimize neural SDF $S_{\boldsymbol{\Theta}_s}$ and diffuse albedo $\beta_{\boldsymbol{\Theta}_\beta}$ by treating $\beta_{\boldsymbol{\Theta}_\beta}$ as a volumetric radiance field. This phase is designed to recover correct object topology and serves as an initialization for the second phase. In the second phase, we refine geometric details and factorize materials from lighting by jointly optimizing neural SDF $S_{\boldsymbol{\Theta}_s}$, neural materials $\beta_{\boldsymbol{\Theta}_\beta},  \kappa_{\boldsymbol{\Theta}_\kappa}$, and $\alpha_{\boldsymbol{\Theta}_\alpha}$, and light intensity $L$ via an edge-aware physics-based surface rendering method.

\subsection{Volumetric radiance field rendering}
In the first stage, we optimize 
$\beta_{\boldsymbol{\Theta}_\beta }(\boldsymbol{x},\boldsymbol{n},\boldsymbol{n},\boldsymbol{f})$ as a view-dependent neural radiance field
(by substituting view direction $-\boldsymbol{d}$ for the second $\boldsymbol{n}$). 
Hence, we perform volumetric radiance field rendering of the neural SDF $S_{\boldsymbol{\Theta}_s}$ and colors $\beta_{\boldsymbol{\Theta}_\beta }(\boldsymbol{x},\boldsymbol{n},\boldsymbol{-d},\boldsymbol{f})$ as in~\cite{wang2021neus} in order to harness the power of volume rendering in recovering correct object topology~\cite{mildenhall2020nerf, oechsle2021unisurf, wang2021neus, yariv2021volume}, i.e., the correct number and location of holes in the geometry.
If we do not perform this initial optimization stage and instead optimize for surface rendering from scratch, we found that the optimization frequently diverges 
unless object segmentation masks are provided, as observed by~\cite{oechsle2021unisurf}, and often gets stuck in local minima with incorrect topology. At the same time, the volumetric nature of this stage is inconsistent with our goal of producing shape and materials that are compatible with the mesh-based rendering paradigm used in existing graphics pipelines. This motivates the second optimization stage where we perform edge-aware physics-based surface rendering.

\algnewcommand\algorithmcinput{\textbf{Input:}}
\algnewcommand\algorithmcoutput{\textbf{Output:}}
\algnewcommand\algorithmchyper{\textbf{Hyperparams:}}
\algnewcommand\INPUT{\item[\algorithmcinput]}
\algnewcommand\OUTPUT{\item[\algorithmcoutput]}
\algnewcommand\HYPER{\item[\algorithmchyper]}

\begin{algorithm}[t]
\caption{Locate edge points}\label{alg:locate_edge_points}
\begin{algorithmic}[1]
\INPUT ray-surface intersection $\hat{\boldsymbol{x}}$ 
\OUTPUT an edge point or NOT\_FOUND.
\HYPER max \# steps $K$, step size $\epsilon$, threshold $\delta$.
\State $\boldsymbol{x}_t \gets \hat{\boldsymbol{x}}$
\For{$i\gets 1$ to $K$}
\If {$(\frac{\boldsymbol{x}_t-\boldsymbol{o}}{\Vert \boldsymbol{x}_t-\boldsymbol{o}\Vert_2})^T\boldsymbol{n}_t<\delta$}
    \State \Return $\boldsymbol{x}_t$
\Else
    \State $\boldsymbol{x}_{t}\gets\boldsymbol{x}_{t}+\epsilon\cdot (\boldsymbol{n}_t-\frac{\boldsymbol{o}-\boldsymbol{x}_t}{(\boldsymbol{o}-\boldsymbol{x}_t)^T\boldsymbol{n}_t})$
\EndIf
\EndFor
\State \Return NOT\_FOUND
\end{algorithmic}
\end{algorithm}

\begin{figure*}[htb]%
    \centering
    \subfloat[Assuming known color, prior neural surface rendering methods~\cite{yariv2020multiview,niemeyer2020differentiable} fail to deform a small sphere to a target large sphere due to lack of edge handling.]{{\includegraphics[width=5.15cm]{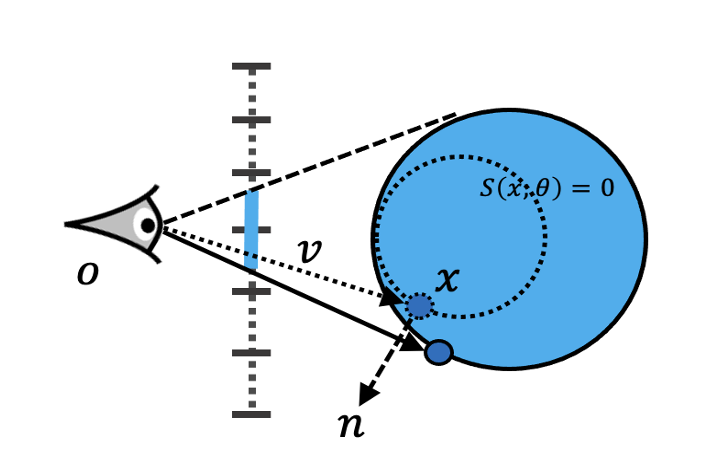} }}%
    \qquad
    \subfloat[The shading at an edge pixel involves the combination of the shading at disconnected surface pieces identified using subpixel localization of edges. ]{{\includegraphics[width=5.15cm]{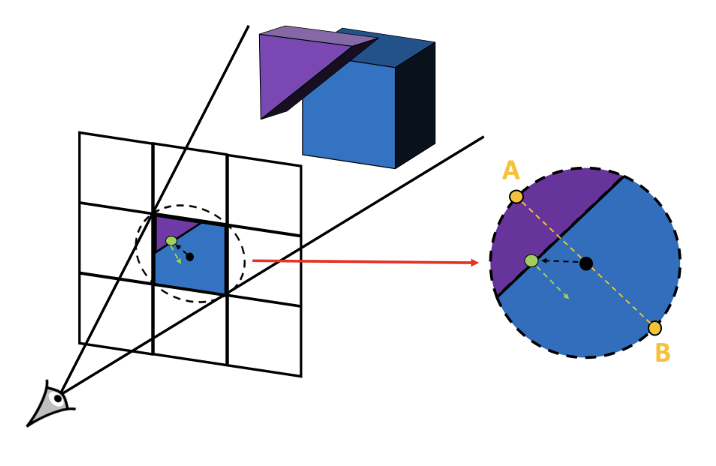} }}%
    \qquad
    \subfloat[We locate edge points in 3D by walking on the zero level set of the neural SDF, then project 3D edge points into 2D for subpixel edge localization.]{{\includegraphics[width=5.15cm]{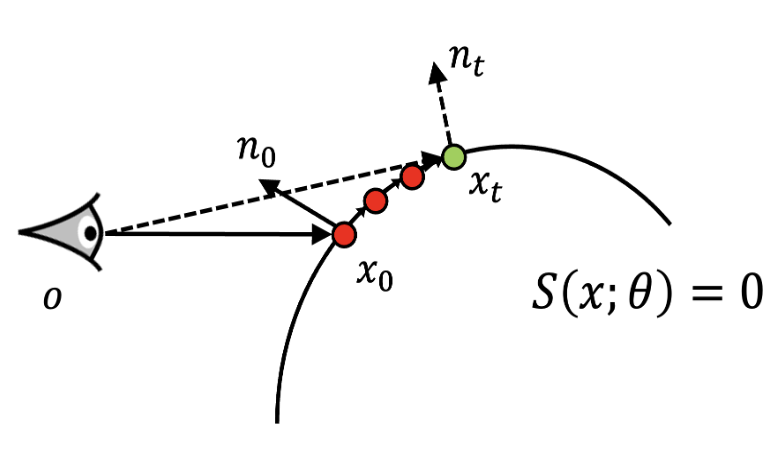} }}%
    \vspace{-0.25\baselineskip}
    \caption{Illustration of edge-aware surface rendering for neural SDFs. (a) Existing neural surface rendering methods ignore geometric discontinuities, making it difficult to deform neural SDFs to match silhouettes even for simple objects. (b) Geometric discontinuities are introduced by edge pixels where multiple depth values are present in a single pixel, motivating our proposed edge sampling algorithm. 
    (c) Our method can localize subpixel-accurate edges for neural SDFs enabling correct shading calculations at edge pixels. }%
    \label{fig:edge_illustrate}%
\end{figure*}

\begin{figure*}[htb]
    \centering
    \includegraphics[width=0.975\textwidth]{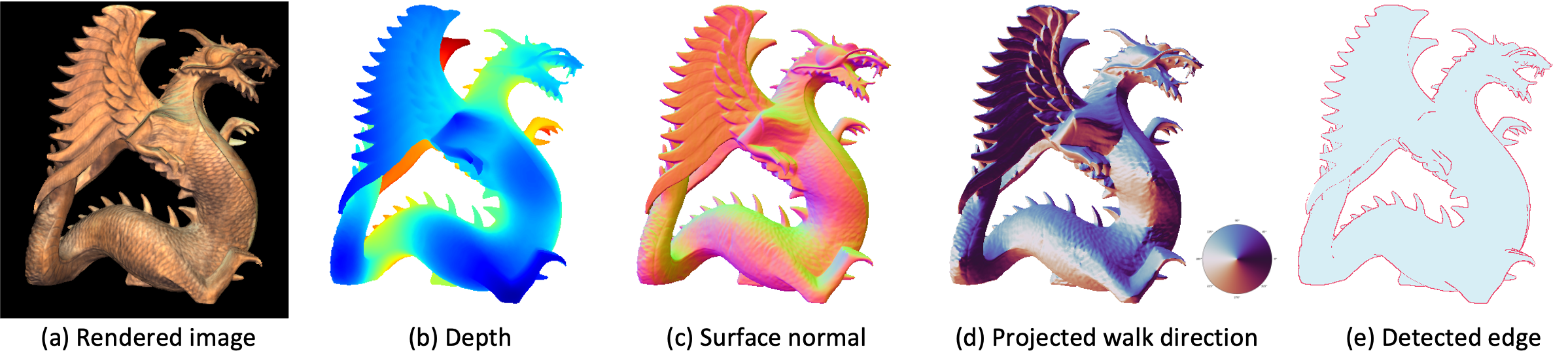}
    \vspace{-0.25\baselineskip}
    \caption{Demonstration of edge-aware surface rendering for neural SDF. To render the image (a), we first run sphere tracing to get the depth (b) and surface normal (c) corresponding to the pixel centers. We then localize subpixel-accurate edges by walking on the surface; the projected walk direction (d) is color-coded as shown in the inset color wheel. Detected edge pixels are shown in red in (e). We shade non-edge pixels as in  existing neural surface rendering methods~\cite{yariv2020multiview, niemeyer2020differentiable}, but shade edge pixels by accounting for the geometric discontinuity.}
    \vspace{-0.25\baselineskip}
    \label{fig:edge_detection_example}%
    \vspace{-0.5\baselineskip}
\end{figure*}

\subsection{Edge-aware physics-based surface rendering}
After solving for a volumetric radiance field, we perform a second, full inverse rendering stage that jointly optimizes neural SDF $S_{\boldsymbol{\Theta}_s}$ and neural materials $\beta_{\boldsymbol{\Theta}_\beta},\kappa_{\boldsymbol{\Theta}_\kappa}$, and $\alpha_{\boldsymbol{\Theta}_\alpha}$) from photometric images.
This stage has two key components: differentiable physics-based shading and edge-aware surface rendering.

\medskip
\noindent \textbf{Physics-based shading.} As our photometric image inputs have co-located flashlight and camera, the light direction is aligned with the view direction across pixel locations. Hence, we can simplify the rendering equation~\cite{kajiya1986rendering} as:
\begin{align}
    L_o(\viewdir, \location)&=\int_{\Omega}  L_i(\lightdir, \location) f_r(\viewdir, \lightdir, \location) (\lightdir \cdot \normaldir)  d\lightdir \\
        &\approx  L_i(\viewdir, \location) f_r(\viewdir, \viewdir, \location) (\viewdir \cdot \normaldir) , \label{eq:rendering_eq}
\end{align}
where $L_o, \location,\normaldir,\lightdir,\viewdir, L_i, f_r$ are observed light, surface location, surface normal, light direction, view direction, incident light and BRDF, respectively. We model the white flashlight as a point light source as in~\cite{nam2018practical, luan2021unified}:
\begin{align}
L_i(\viewdir;\location)=\frac{L}{\Vert \location-\boldsymbol{o}\Vert_2^2}, \label{eq:light_source}
\end{align}
where $L$ is a scalar light intensity and $\boldsymbol{o}$ is the light location (same as the camera location in our co-located capture setup). The denominator models the inverse-square fall-off in intensity.
For BRDF $f_r$, we use the GGX model~\cite{walter2007microfacet}, whose parameters at location $\boldsymbol{x}$, diffuse albedo $\beta$, specular albedo $\kappa$ and roughness $\alpha$, are encoded in our neural materials $\beta_{\boldsymbol{\Theta}_\beta},\kappa_{\boldsymbol{\Theta}_\kappa}$, and $\alpha_{\boldsymbol{\Theta}_\alpha}$, respectively.

From Eqns.~\ref{eq:rendering_eq} and \ref{eq:light_source}, we observe that gradients from the rendered image $L_o(\viewdir; \location)$ must back-propagate through surface location $\boldsymbol{x}$ and surface normal $\boldsymbol{n}$ to the shape parameters, and through the BRDF $f_r$ to the material parameters. 

\begin{figure}[t]
    \centering
    \includegraphics[width=0.94\columnwidth]{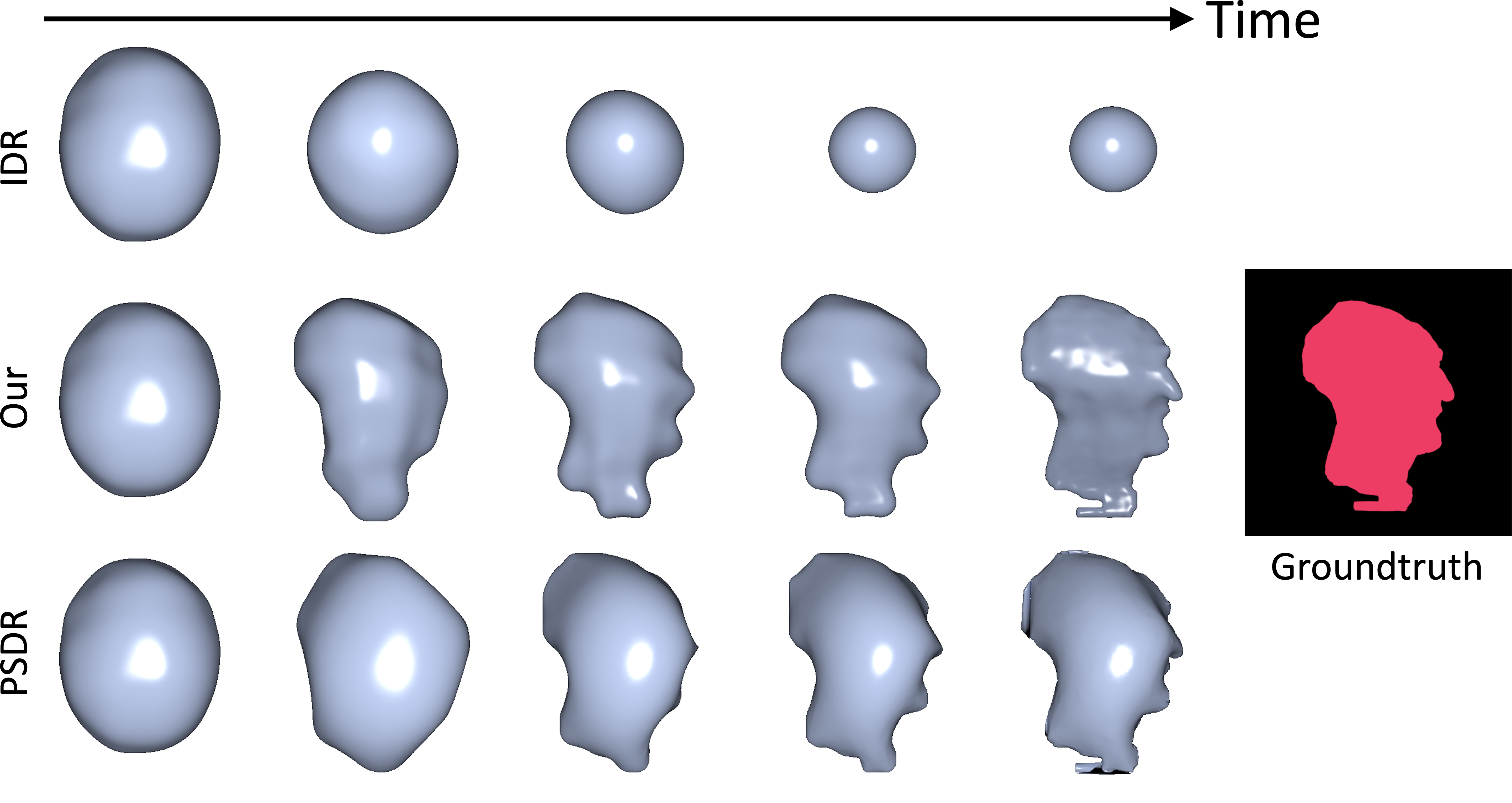}\\
    \vspace{-2mm}
    \caption{Given a single target image of an object with known color, we use an image loss to optimize a neural SDF with both IDR (top row)~\cite{yariv2020multiview} and out IRON (middle row) method. 
    We also optimize a mesh using PSDR (bottom row)~\cite{luan2021unified}. Results show that IDR gets stuck due to not allowing SDF to deform along the image plane, while our method correctly optimizes the sihouette. PSDR can also handle the silhouettes, but the mesh quality degrades without intermediate remeshing steps.  
    }
    \label{fig:single_view}
    \vspace{-2mm}
\end{figure}

\medskip
\noindent \textbf{Edge-aware surface rendering.} Inspired by recent advances in mesh-based differentiable rendering~\cite{li2018differentiable, nimier2019mitsuba, bangaru2020warpedsampling, luan2021unified, Zhang:2021:PSDR}, we identify a key issue in prior neural surface rendering works such as IDR~\cite{yariv2020multiview} and DVR~\cite{niemeyer2020differentiable}. Namely, their differentiable 
rendering module only works for interior pixels, due to their 
assumption of a smooth surface inside each pixel footprint. This assumption fails 
for 
edge pixels, where shading color is a combination of colors at disconnected surface pieces, as shown in Fig.~\ref{fig:edge_illustrate}(b). For this reason, these methods compute biased gradients w.r.t.\ the weights of the neural SDF that 
move surface points along camera rays, but not in the image plane, due to missing edge gradients reflecting how 
color changes w.r.t.\ edge location. In Fig.~\ref{fig:edge_illustrate}(a), we illustrate how these missing edge gradients 
lead to non-convergence in the simple task of deforming an initial small sphere to a target large sphere assuming known shading color. This problem is confirmed empirically in Fig.~\ref{fig:single_view}.

We address this issue 
via a novel edge sampling algorithm tailored for neural SDFs. %
 Our algorithm has three steps:
1) localize subpixel-accurate 2D edges by detecting 3D edge points that are then projected to the 2D image, 2) re-parametrize edge points such that they can back-propagate gradients to the neural SDF in an auto-differentation framework, and 3) compute the shading color for edge pixels.

In step 1, we start from ray-surface intersections found by sphere tracing the center rays at each pixel location~\cite{spheretracing} (Fig.~\ref{fig:edge_detection_example}(b,c)), and walk on the surface along the direction defined in line 6 of Alg.~\ref{alg:locate_edge_points} and illustrated in Fig.~\ref{fig:edge_illustrate}(c) %
until reaching a 3D edge point or a max number of steps~\cite{Bremer98rapidapproximate,Bnard2019LineDF}. We visualize projected walk directions in Fig.~\ref{fig:edge_detection_example}(d). For the sake of efficiency, we only do the surface walk process for the ray-surface intersections at depth discontinuity pixels in order to reduce the number of evaluations of the neural SDF; we identify such depth discontinuity pixels as ones with depth Sobel gradient magnitude above a certain threshold $\tau$.
We then project detected 3D edge points to image space, producing both subpixel-accurate edge locations and an edge mask marking pixels containing edges 
(Fig.~\ref{fig:edge_detection_example}(e)).
We also obtain 2D edge normal directions by projecting the 3D surface normals at these edge points to 2D. 

In step 2, we reparameterize the edge points' locations $\boldsymbol{x}$ to make them differentiable with respect to network weights of the neural SDF. We observe that the differentiable ray-surface intersection equation (Eq.~\ref{eq:interior_reparam}) for interior points in \cite{yariv2020multiview,niemeyer2020differentiable} only captures how perturbations to neural SDF weights move the ray-surface intersection along the camera ray, while for edge points, we care about their movement along the surface normal. Hence, to reparameterize edge points correctly,  we propose to replace the viewing direction $\boldsymbol{o}-\boldsymbol{x}$ ($\boldsymbol{o}$ is the camera origin, $\boldsymbol{x}$ is a surface point) in Eq.~\ref{eq:interior_reparam} with the surface normal $\boldsymbol{n}$, and arrive at Eq.~\ref{eq:edge_reparam}. 
\begin{align}\label{eq:interior_reparam}
    \boldsymbol{x}_{\boldsymbol{\Theta}_s}=\boldsymbol{x}-\frac{\boldsymbol{\boldsymbol{o}-\boldsymbol{x}}}{\boldsymbol{n}^T(\boldsymbol{o}-\boldsymbol{x})}S_{\boldsymbol{\Theta}_s}(\boldsymbol{x}),
\end{align}
\begin{align}\label{eq:edge_reparam}
   \boldsymbol{x}_{\boldsymbol{\Theta}_s}=\boldsymbol{x}-\frac{\boldsymbol{n}}{\boldsymbol{n}^T\boldsymbol{n}}S_{\boldsymbol{\Theta}_s}(\boldsymbol{x})=\boldsymbol{x}-\boldsymbol{n}S_{\boldsymbol{\Theta}_s}(\boldsymbol{x}).
\end{align}
We show the correctness of our edge point reparametrization in the supplemental material. 

In step 3, we compute the shading at each edge pixel. Consider the edge pixel in Fig.~\ref{fig:edge_illustrate}(b), and let the green projected edge point have subpixel coordinates $[u, v]$, and the green projected surface normal be $[du, dv]$. Then the black center of this edge pixel has coordinates:
\begin{align}
    [u_c, v_c] = \big[\mathrm{floor}(u), \mathrm{floor}(v)\big] + 0.5.
\end{align}
We approximate each square pixel footprint using a circle of radius $\frac{\sqrt{2}}{2}$ pixels centered at $[u_c, v_c]$. We then pick 2D locations, labeled A and B, on the circle on either side of the edge.
We raytrace and shade the two selected 2D locations. Let us denote the shaded colors as $C_A$ and $C_B$, respectively. We linearly combine $C_A$ and $C_B$ with weights proportional to the two segment areas separated by the edge. Suppose the fraction of segment area on the same side as A is $w_A\in [0, 1]$:
\begin{align}
    \alpha &=2\cdot\arccos({\sqrt{2}\cdot[du, dv][u-u_c, v-v_c]^T}), \label{eq:edge_combine_weight_1}\\
    w_A&=1-\frac{1}{2\pi}\cdot \big(\alpha-\sin\alpha\big),\label{eq:edge_combine_weight_2}
\end{align}
then our predicted edge pixel color is:
\begin{align}
C=w_AC_A+(1-w_A)C_B.    \label{eq:edge_pixel_color}
\end{align}
In Eq.~\ref{eq:edge_pixel_color}, gradients from $C$ can back-propagate through $w_A, C_A$, and $C_B$ to the parameters of our neural SDF and materials, and through the variable $w_A$, we correctly model how tiny perturbations of the 3D edge point's location along the surface normal direction affect the edge pixel color.

\begin{figure}[t]
    \centering
    \includegraphics[width=0.85\columnwidth]{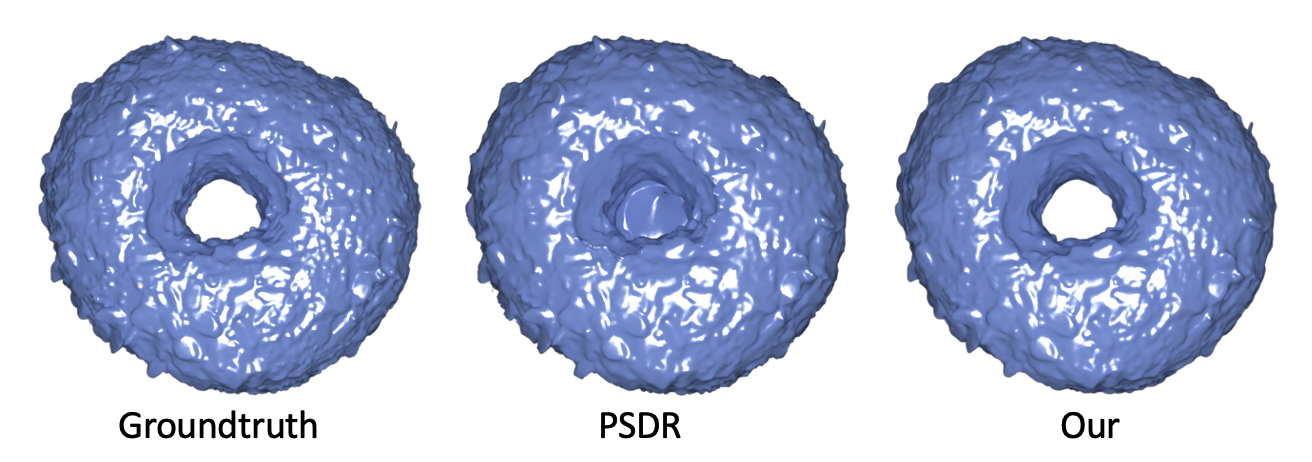}\\
    \vspace{-2mm}
    \caption{When using a mesh representation, it is difficult to design methods that can change the mesh topology in a differentiable way. In contrast, methods that use neural SDFs can more readily change their topology. Above, we use the mesh-based differentiable rendering pipeline PSDR~\cite{Zhang:2021:PSDR} and our IRON pipeline to reconstruct this bagel by deforming an initial sphere. The results demonstrate that PSDR fails to recover the central hole, while our method has no such issue. 
    }
    \label{fig:bagel_fig}
    \vspace{-2mm}
\end{figure}

\subsection{Training and testing}
Given multi-view photometric images, we optimize our neural SDF and materials using the following loss:
\begin{align}
L=&L_2\big(\text{pyramid}(\hat{I}),\text{pyramid}(I)\big) \label{eq:l2loss}\\
  &+ 1- \text{SSIM}(\hat{I}, I) \label{eq:ssimloss}\\
  &+\lambda_1\cdot \Vert\nabla_x S - 1\Vert_2^2 \label{eq:eikloss}\\
  &+\lambda_2\cdot \max(\text{roughness}(x)-0.5, 0), \label{eq:roughrangeloss}
\end{align}
where Eq.~\ref{eq:l2loss} is the $L_2$ loss on Gaussian pyramids of the predicted image $\hat{I}$ and groundtruth image $I$, Eq.~\ref{eq:ssimloss} is the SSIM loss~\cite{ssim_paper}, Eq.~\ref{eq:eikloss} is the eikonal loss~\cite{gropp2020implicit, crandall1983viscosity} enforcing the validity of the SDF, and Eq.~\ref{eq:roughrangeloss} is the roughness range loss encouraging the estimated roughness to stay below 0.5. $\lambda_1, \lambda_2$ are loss weights.  

Once training is complete, we convert the neural SDF and materials to a triangle mesh and texture map for deployment in the standard graphics pipeline. We first extract a mesh from the optimized neural SDF using the marching cube algorithm~\cite{lorensen1987marching}.
We then use the Blender Smart UV Project tool~\cite{blender} to compute a reasonable per-vertex uv mapping.
Finally, to fill the material texture images, we densely sample points on our triangle meshes with trilinearly interpolated uv coordinates, then query the material networks at the sampled surface points, and finally splat the per-point material parameters to the texture images using their interpolated uv coordinates.

\begin{figure*}[htp]
	\centering
    \includegraphics[width=0.95\textwidth]{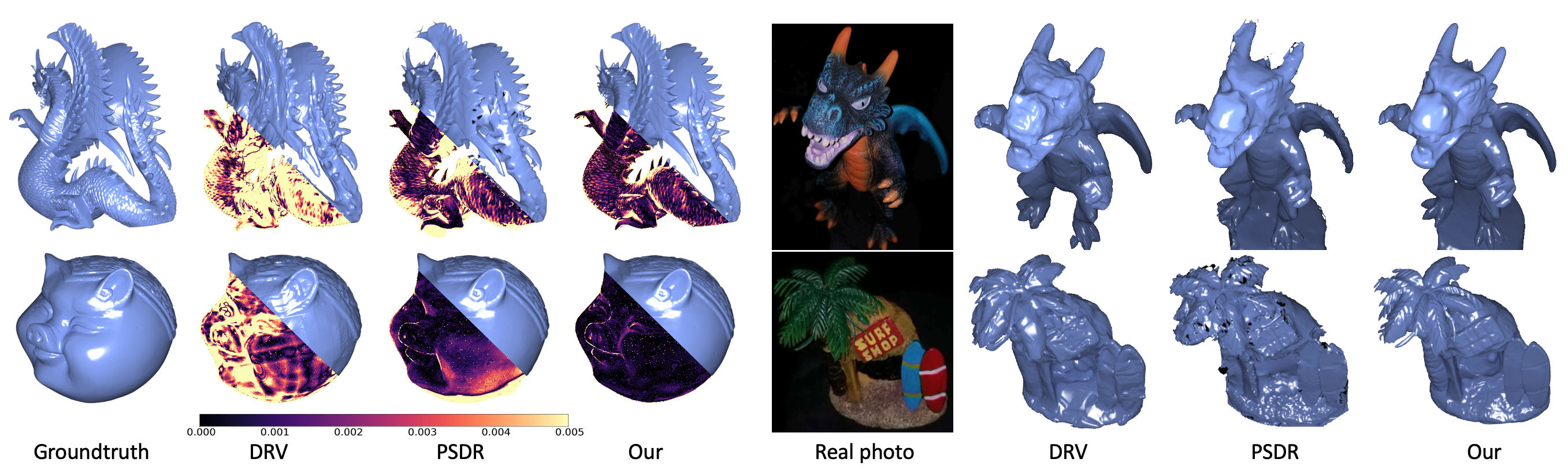}
    \vspace{-0.25\baselineskip}
    \caption{Comparison of recovered geometry on synthetic (\textbf{left}) and real data (\textbf{right}) with baselines DRV~\cite{bi2020deepreflectance} and PSDR~\cite{luan2021unified}. On synthetic data, we achieve the highest geometric reconstruction accuracy among the inverse rendering methods designed for flashlight photography. For real data, we show a captured photo for reference due to lack of ground truth geometry. }
    \label{fig:mesh_cmp}
    \vspace{-0.25\baselineskip}
\end{figure*}

\begin{figure*}[t]
	\centering
     \includegraphics[width=0.95\textwidth]{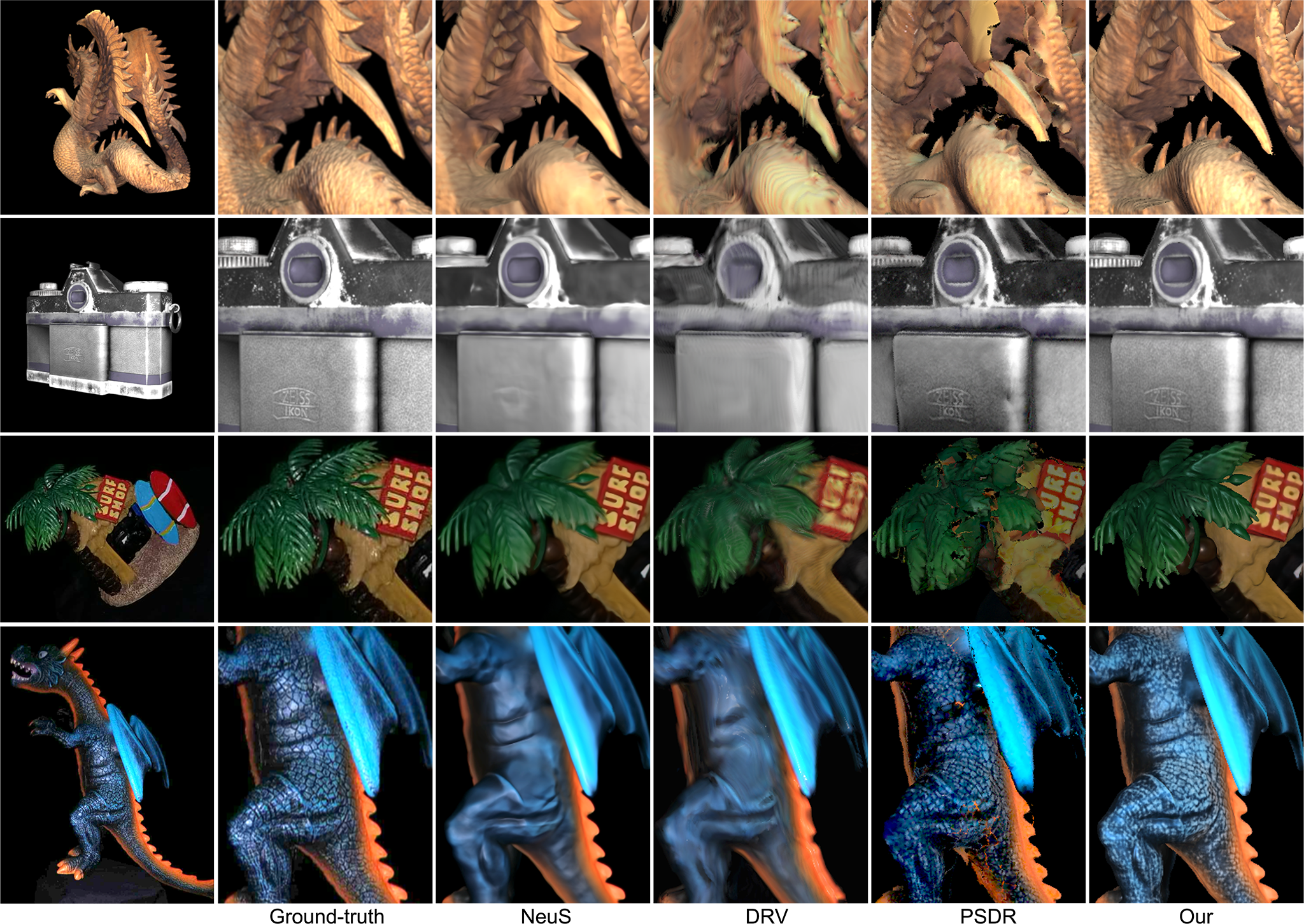}   
    \vspace{-0.25\baselineskip}
    \caption{Qualitative comparison of generalization to novel co-located flashlight relighting using both synthetic (\textbf{top two rows}) and real (\textbf{bottom two rows}) data. Our IRON method produces visually more faithful matches in terms of geometric and texture details compared to DRV~\cite{bi2020deepreflectance} and PSDR~\cite{luan2021unified}. Results of the recent neural surface reconstruction work NeuS~\cite{wang2021neus} are also included for reference.}
    \vspace{-0.75\baselineskip}
    \label{fig:novel_colocated}
\end{figure*}

\begin{figure*}[t]
	\centering
    \includegraphics[width=0.96\textwidth]{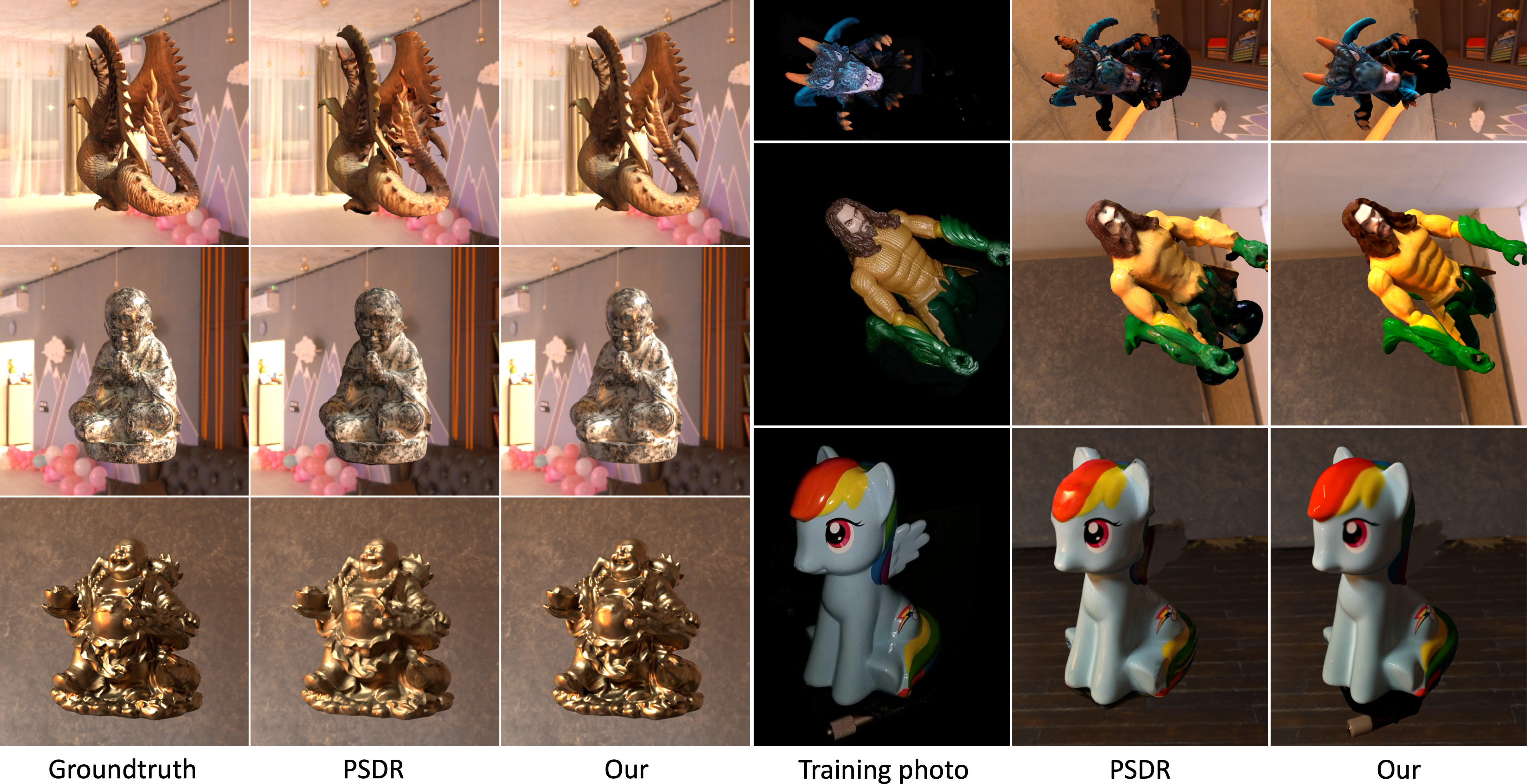}
    \vspace{-0.25\baselineskip}
    \caption{Comparison of generalization to novel natural environmental lighting on both synthetic (\textbf{left}) and real (\textbf{right}) data. We render both the PSDR~\cite{luan2021unified} and our reconstructions under environmental illuminations using Mitusba~\cite{Mitsuba}. Note that our IRON pipeline matches the ground truth better for specular highlight regions on synthetic data, and is more perceptually convincing on real data.
    }
    \label{fig:novel_envmap}
\end{figure*}

\begin{table}
\centering
\resizebox{\columnwidth}{!}{
\begin{tabular}{l@{\hspace{1.5em}} c@{\hspace{1em}}@{\hspace{0.5em}}c@{\hspace{1em}}c@{\hspace{1em}}c@{\hspace{1em}}@{\hspace{0.5em}}} 
\toprule 
&  $\downarrow$Chamfer L1~\cite{aanaes2016large} & $\downarrow$LPIPS~\cite{zhang2018perceptual} & $\uparrow$SSIM~\cite{ssim_paper} & $\uparrow$PSNR \\
\midrule
DRV~\cite{bi2020deepreflectance}  & 0.0111 & 0.1133 & 0.8252 & 28.0693 \\
PSDR~\cite{luan2021unified} & 0.0048   & 0.1032 & 0.9358  &27.1354 \\
Our & \textbf{0.0014} & \textbf{0.0438} & \textbf{0.9747} & \textbf{31.2614}  \\
\bottomrule
\end{tabular}
}
\vspace{-0.25\baselineskip}
\caption{Quantitative comparisons with baseline methods on synthetic data in terms of geometry quality and relighting quality under novel co-located flashlight illumination. Note all synthetic objects are scaled to lie inside the unit sphere before evaluation.}\label{tab:compare_synthetic}
\vspace{-0.25\baselineskip}
\end{table}

\begin{table}
\centering
\resizebox{0.726\columnwidth}{!}{
\begin{tabular}{l@{\hspace{1.5em}} c@{\hspace{1em}}c@{\hspace{1em}}c@{\hspace{1em}}@{\hspace{0.5em}}} 
\toprule 
& $\downarrow$LPIPS~\cite{zhang2018perceptual} & $\uparrow$SSIM~\cite{ssim_paper} & $\uparrow$PSNR \\
\midrule
DRV~\cite{bi2020deepreflectance}   & \textbf{0.1016} & 0.8264 & \textbf{32.0303}		  \\
PSDR~\cite{luan2021unified}  & 0.1861 &	0.8137	 & 25.7452 \\
Our &	0.1091 & \textbf{0.8614} & 29.3694 \\
\bottomrule
\end{tabular}
}
\vspace{-0.25\baselineskip}
\caption{Quantitative comparisons on real data in terms of relighting quality under novel co-located flashlight illumination.}\label{tab:compare_real}
\vspace{-0.25\baselineskip}
\end{table}

\section{Evaluation}
We perform extensive experiments to validate IRON.

\subsection{Optimizing neural SDFs to fit single image}
We design a controlled experiment to illustrate the key effect of 
our edge sampling algorithm: allowing the image loss to lead to in-plane neural SDF deformations. In this experiment, we assume that a single image of a 3D object with known constant color is given as input to optimize a neural SDF through an image loss. To solve this problem, the neural SDF must be deformed such that its silhouette matches the ground truth when viewed from the input viewpoint. But as shown in Fig.~\ref{fig:single_view}, the prior neural surface rendering method IDR~\cite{yariv2020multiview} fails to accomplish this task due to a lack of edge pixel handling. 
Our 
edge-aware surface rendering method addresses this problem, converging to a valid solution. 
Although mesh-based differentiable rendering methods can also handle silhouettes~\cite{Zhang:2021:PSDR}, we observe degraded mesh quality over the course of optimization without intermediate remeshing, due to the otherwise fixed mesh topology.
In contrast, neural SDFs
do not suffer from such problems.

\subsection{Inverse rendering from photometric images}
We now evaluate methods on the task of inverse rendering from multi-view photometric images.

\medskip
\noindent \textbf{Datasets.} We create a synthetic dataset consisting of 9 
objects: \emph{dragon}, \emph{buddha}, \emph{camera}, \emph{monk}, \emph{kettle}, \emph{duck}, \emph{pig}, \emph{sneaker}, and \emph{bagel}. 
Each object is rendered from 200 randomly sampled viewpoints using the Mitsuba path-tracing renderer~\cite{Mitsuba} to form the training data. We co-locate a point light source with the camera to actively illuminate the object without other light sources. We render test data consisting of 100 images under novel co-located flashlight illumination, and another 100 images under novel natural environmental illumination. 
For real-world data, we use 5 object captures from DRV~\cite{bi2020deep}: \emph{dragon}, \emph{pony}, \emph{girl}, \emph{tree}, and \emph{triton}. The data is acquired using co-located flashlight setup in a dark environment. We randomly choose 70\% of the images for training and use the remaining 30\% as test images for evaluating 
generalization to novel view with co-located lighting.

\medskip
\noindent \textbf{Baselines.} 
We compare with two 
inverse rendering methods that use
densely captured photometric images: the volume-based DRV~\cite{bi2020deep} and the mesh-based PSDR~\cite{luan2021unified} methods. We did not compare with the mesh-based method of Nam \etal~\cite{nam2018practical} due to the lack of open-source code. In addition, PSDR has shown superior reconstruction quality compared to this method.

\medskip
\noindent \textbf{Discussion.} As shown in Tab.~\ref{tab:compare_synthetic} and
Figs.~\ref{fig:mesh_cmp} and~\ref{fig:novel_colocated}, we outperform state-of-the-art inverse rendering baselines by a large margin in terms of geometric accuracy and generalization to novel co-located lighting on synthetic data, and produce fewer artifacts like blurry textures on real data. 
The mesh-based PSDR method sometimes produces
geometry with incorrect topology (number of holes), as shown by the bagel data in Fig.~\ref{fig:bagel_fig}.
Such mesh-based methods also face difficulties in maintaining mesh regularity and avoiding self-intersections
during optimization, and
require repeated application of error-prone remeshing and uv-mapping procedures. 
Neural SDFs avoid these problems with their continuous representation. We also compare with PSDR in terms of generalization to novel environmental lighting in Fig.~\ref{fig:novel_envmap}. IRON results in better synthesized specular highlights and more perceptually convincing relighting. 
The volumetric DRV results are not directly renderable by Mitsuba; hence we did not compare with them under novel environmental lighting. 
On real data under novel co-located flashlight relighting, IRON's material estimates 
are much more detailed than those of baseline methods (Fig.~\ref{fig:novel_colocated}), with quantitative metrics on par with DRV (Tab.~\ref{tab:compare_real}). While DRV has slightly better LPIPS and PSNR scores, we observe that it significantly blurs detail. In addition, our method enjoys the advantage of simple conversion to a mesh for computer graphics deployment.

\begin{figure}[t] 
    \centering
    \includegraphics[width=0.99\columnwidth]{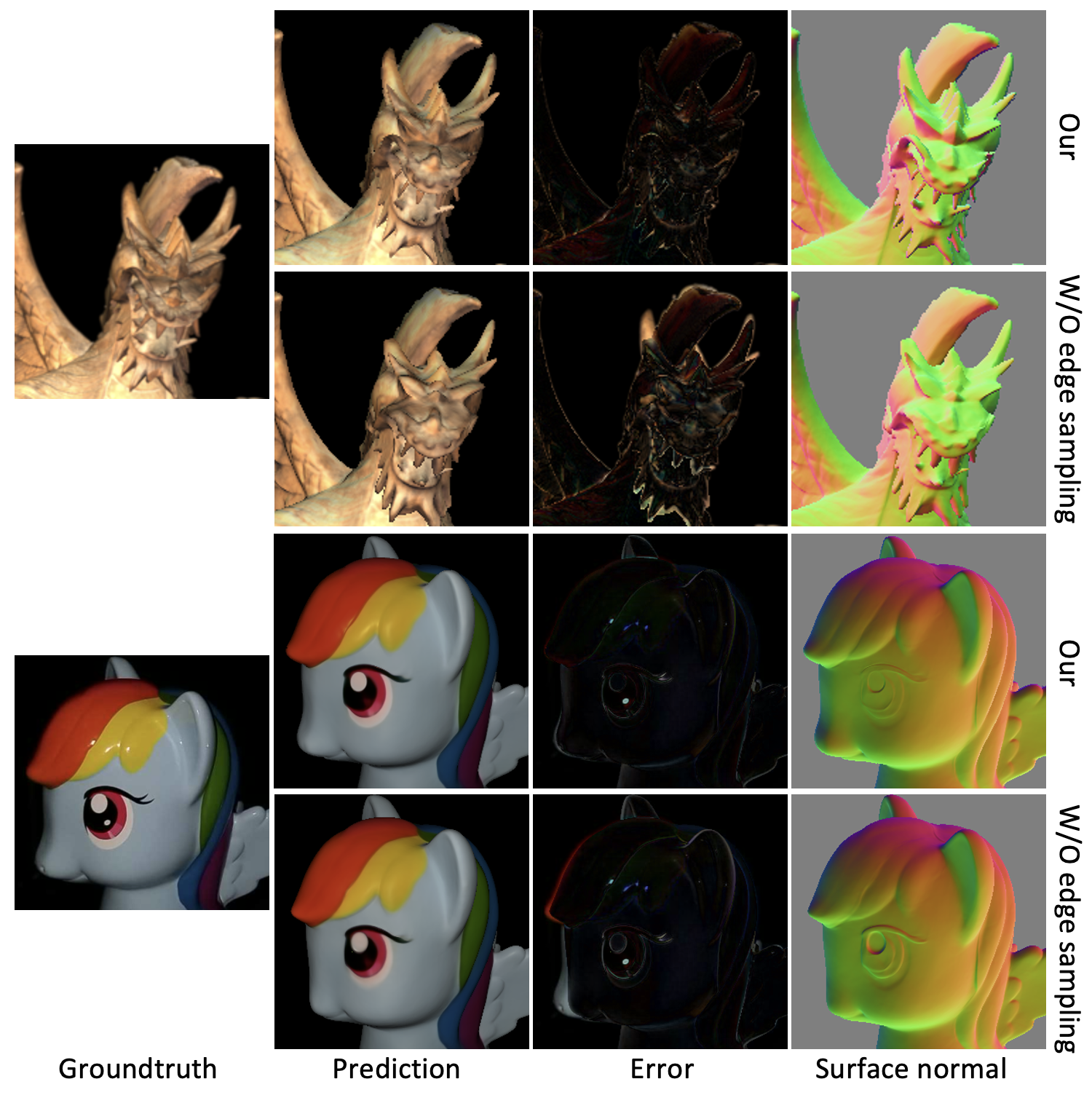}
    \vspace{-2mm}
    \caption{Ablations on our edge sampling algorithm using the synthetic dragon and real pony data. Without edge sampling, neural surface rendering performs poorly around edge regions, as shown by the error maps and surface normals.  }
    \vspace{-2mm}
    \label{fig:ablation_edge}
\end{figure}

\medskip
\noindent \textbf{Ablations on edge sampling.} We test removing the 
our 
edge sampling method from IRON. As shown in Fig.~\ref{fig:ablation_edge}, the resulting reconstructions have much poorer quality around edge regions, e.g., the dragon horn and pony nose.
This poor quality is 
due to the missing edge derivatives in existing neural surface rendering methods~\cite{yariv2020multiview,niemeyer2020differentiable}.

\begin{figure}[t]
    \centering
    \includegraphics[width=0.99\columnwidth]{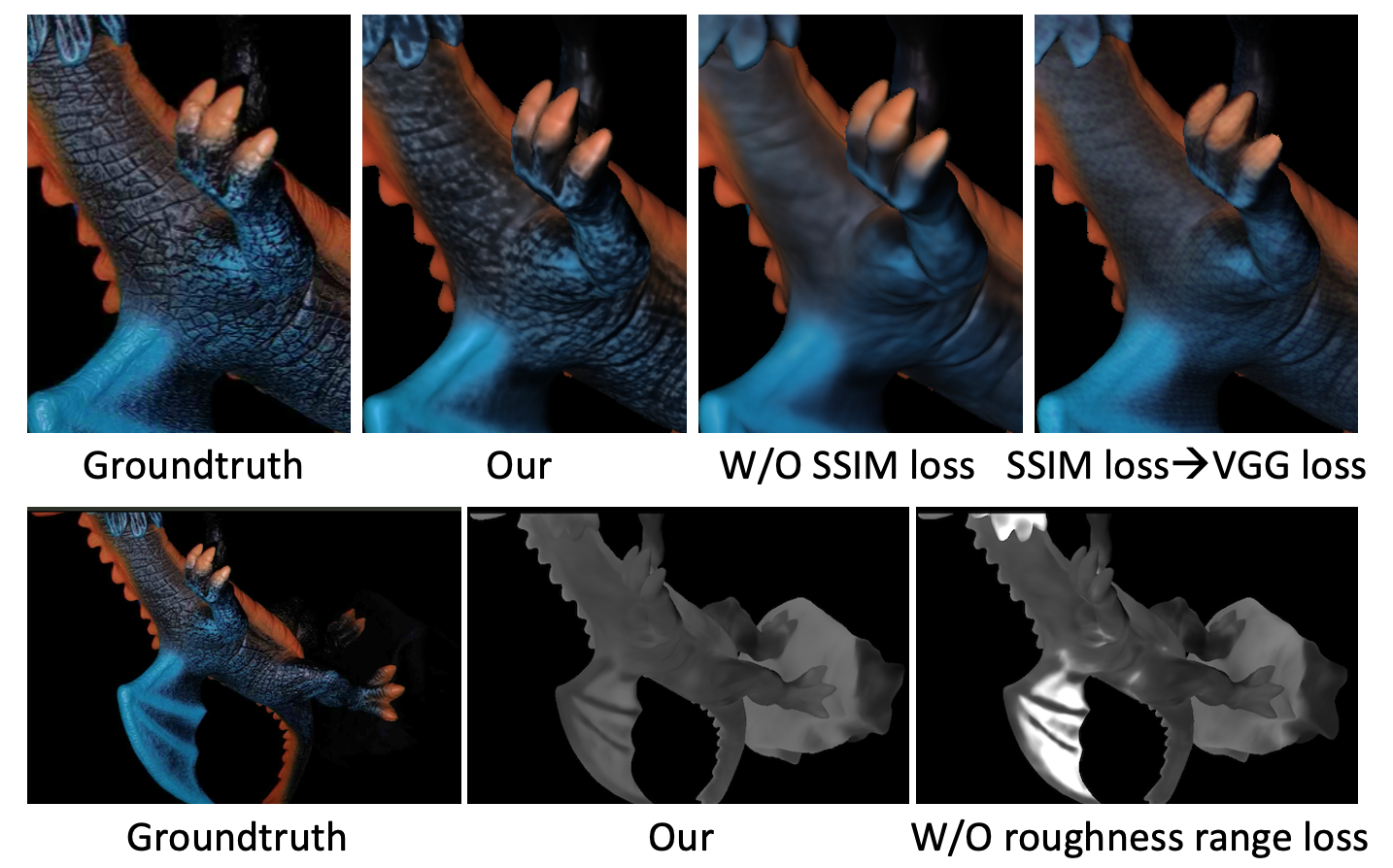}
    \vspace{-2mm}
    \caption{Ablations on our loss functions using the real dragon data. Without patch-based SSIM loss, our synthesized image is quite blurry, while VGG loss results in spurious texture details. The roughness range loss fixes issue of over-estimating roughness.
    }
    \label{fig:ablation_loss}
    \vspace{-1mm}
\end{figure}

\medskip
\noindent \textbf{Ablations on loss functions.} We find that using the SSIM loss significantly improves the sharpness of reconstructions. On the real dragon data, we first try removing SSIM loss
from our pipeline. Then we also try replacing SSIM loss with VGG loss~\cite{Johnson2016Perceptual}, which is widely used in view synthesis tasks. As shown in Fig.~\ref{fig:ablation_loss}, our proposed loss functions lead to the sharpest and most plausible material reconstructions. %

\section{Conclusion}
In this work, we have presented our new IRON pipeline specialized in high-quality inverse rendering from photometric images. In constrast to mesh-based differentible rendering, IRON adopts neural SDFs and materials as representations for ease of optimization in the form of both volumetric radiance rendering and edge-aware physics-based surface rendering, and preserves the convertibility to meshes and material textures for the sake of downstream applications.  

\medskip 
\noindent \textbf{Limitations and future work.} 
There are limitations to be resolved in future work. First, the use of photometric images leads to a more involved data capture process, although such images simplify inverse methods because of the known single point light and minimal shadows. Future work can explore extensions of our work to the combination of flashlight and ambient illumination. Second, we do not model multiple bounces of light. This can lead to material estimation errors in concave regions that feature significant interreflection. Future directions include devising computationally-efficient global illumination rendering algorithms for neural SDF representations. Third, our current BRDF model assumes opaque surfaces, and thus we do not expect our method to work well on transparent and translucent objects that feature significant refraction and subsurface scattering.

\medskip
\noindent \textbf{Acknowledgements.} This work was supported in part by the National Science Foundation (IIS-2008313), and by funding from Intel and Amazon Web Services. We also thank Sai Bi for providing their code and data.

\appendix

\begin{figure*}[ht]
    \centering
    \includegraphics[width=0.998\textwidth]{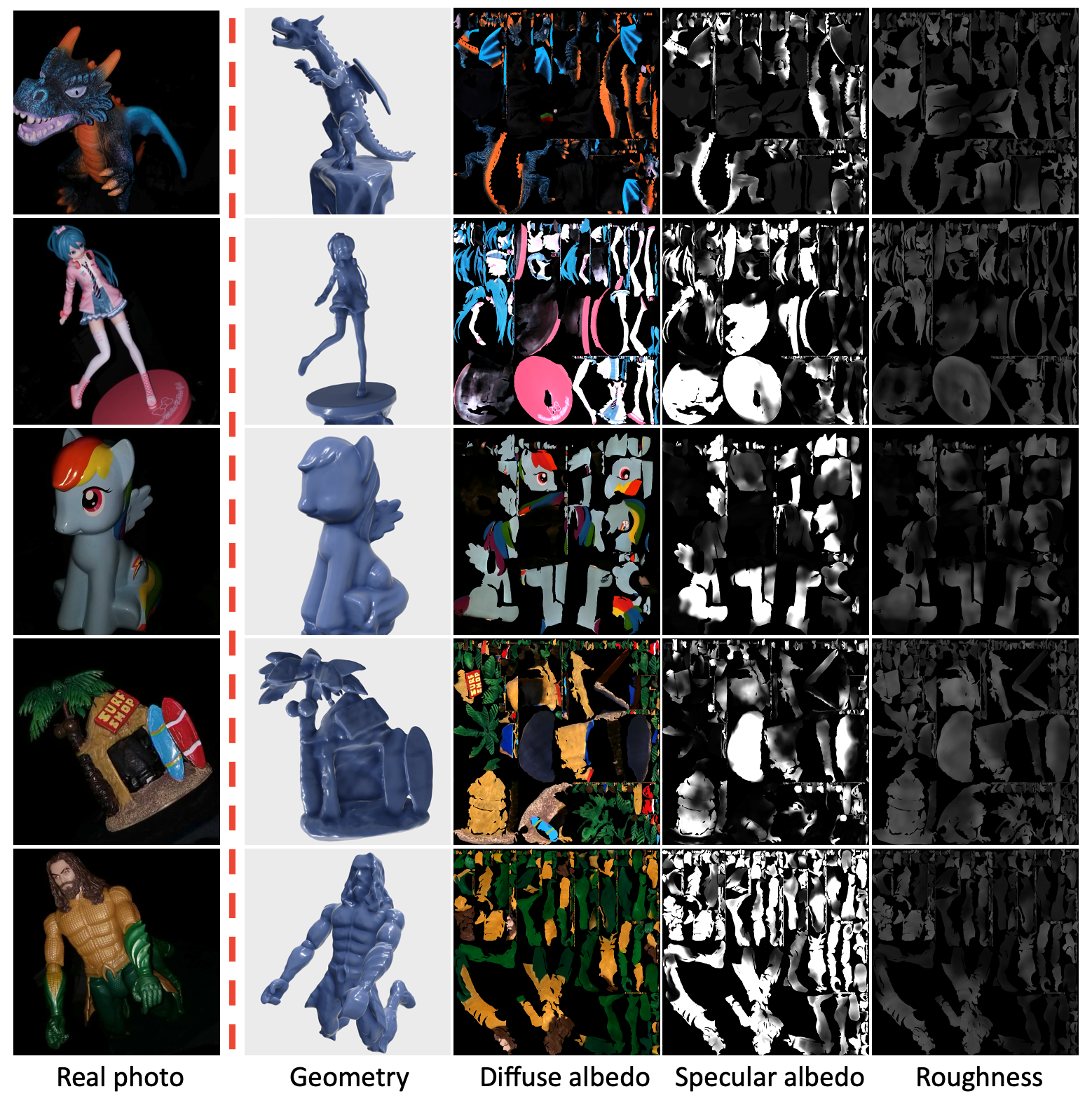}
    \caption{Reconstructed meshes and materials for the real scenes by our IRON system. }
    \label{fig:assets}%
\end{figure*}

\section{Neural network structures}
\noindent\textbf{Neural SDF} $S_{\boldsymbol{\Theta}_s}: \boldsymbol{x} \xrightarrow{}(S, \boldsymbol{f})$. We use an 8-layer MLP of width 256 and a skip connection at the 4th layer. The input 3D location $\boldsymbol{x}$ is encoded by positional encoding using 6 frequencies to compensate the spectral bias of MLPs~\cite{mildenhall2020nerf,tancik2020fourier}. 

\smallskip
\noindent\textbf{Neural diffuse albedo} $\beta_{\boldsymbol{\Theta}_\beta}: (\boldsymbol{x},\boldsymbol{n},\boldsymbol{n},\boldsymbol{f}) \xrightarrow{}\beta$. We use a 8-layer MLP of width 256 and a skip connection at the 4th layer. The input 3D location $\boldsymbol{x}$ is positional-encoded using 10 frequencies. The second surface normal $\boldsymbol{n}$ (equals the viewing direction in the first stage of volumetric radiance fields rendering) is positional-encoded using 4 frequencies.

\smallskip
\noindent\textbf{Neural specular albedo} $\kappa_{\boldsymbol{\Theta}_\kappa}: (\boldsymbol{x},\boldsymbol{n},\boldsymbol{f})\xrightarrow{}\kappa$. We use a 4-layer MLP of width 256, with input 3D location $\boldsymbol{x}$ positional-encoded using 6 frequencies. 

\smallskip
\noindent\textbf{Neural roughness} $\alpha_{\boldsymbol{\Theta}_\alpha}: (\boldsymbol{x},\boldsymbol{n},\boldsymbol{f})\xrightarrow{}\alpha$. We use a 4-layer MLP of width 256, with input 3D location $\boldsymbol{x}$ positional-encoded using 6 frequencies.

\section{Implementation details}
We implement our BRDF by following the Mitsuba roughplastic BRDF implementation~\cite{Mitsuba} closely. The \emph{distribution} parameter in Mitsuba roughplastic BRDF is chosen as ``ggx", while the \emph{intIOR}, \emph{extIOR}, and \emph{nonlinear} parameters are set to their default values. 

During the volumetric radiance field rendering optimization stage, we train for $100k$ iterations using $512$ randomly sampled pixels at each iteration with $\ell_1$ image loss and the eikonal regularization loss (weight $\lambda_1=0.1$). During the edge-aware physics-based surface rendering stage, we set the eikonal loss weight $\lambda_1=0.1$, and the roughness range loss weight $\lambda_2=0.1$. We set $\tau=\expnumber{1}{-2}$ for the depth gradient magnitude threshold, $K=16$ for the maximum number of surface walk steps, $\epsilon=\expnumber{1}{-3}$ for the step size, and $\delta=\expnumber{5}{-2}$ for the dot-product threshold when localizing edge points. 
At each training iteration, we render a random $128\times 128$ image patch to compare with ground truth. The number of Gaussian pyramid levels for $\ell_2$ image loss is set to 4. We train the second stage for $50k$ iterations. The two stages take $\sim$10 hours on a single NVIDIA RTX2080Ti GPU with 12G memory.

\section{Proof of edge point re-parametrization}
In this section, we prove the correctness of our edge point re-parametrization in Eq. (4) of the main paper. Our proof strategy is similar to the differentiable ray-surface intersection in~\cite{yariv2020multiview, niemeyer2020differentiable}, and divided into 3 steps: 1) re-parametrize the edge point in terms of the target moving direction, then show both 2) the function value and 3) the first derivative value under current parameter setting.

First, we note that our goal is to move a 3D edge point $\boldsymbol{x}$ along its surface normal direction $\boldsymbol{n}$; hence:
\begin{align}
    \boldsymbol{x}_{\boldsymbol{\Theta}_s}=\boldsymbol{x}+t_{\boldsymbol{\Theta}_s}\cdot \boldsymbol{n}. \label{eq:parametrize}
\end{align}

Second, we note that under current parameter setting, i.e., $\boldsymbol{\Theta}_s=\boldsymbol{\Theta}_s^{(0)}$, we have:
\begin{align}
    \boldsymbol{x}_{\boldsymbol{\Theta}_s}\bigg\rvert_{\boldsymbol{\Theta}_s=\boldsymbol{\Theta}_s^{(0)}}=\boldsymbol{x}.\label{eq:edge_point_value}
\end{align}
This is to say that the function $\boldsymbol{x}_{\boldsymbol{\Theta}_s}$ evaluated at $\boldsymbol{\Theta}_s^{(0)}$ equals the edge point location $\boldsymbol{x}$. Substituting Eq.~\ref{eq:parametrize} into Eq.~\ref{eq:edge_point_value} gives:
\begin{align}
t_{\boldsymbol{\Theta}_s}\bigg\rvert_{\boldsymbol{\Theta}_s=\boldsymbol{\Theta}_s^{(0)}}=0.\label{eq:zero_order}
\end{align}

Third, we note that $\boldsymbol{x}_{\boldsymbol{\Theta}_s}$ must stay on the zero level set during deformation:
\begin{align}
    S_{\boldsymbol{\Theta}_s}(\boldsymbol{x}_{\boldsymbol{\Theta}_s})=0.
\end{align}
Implicit-differentiate with respect to $\boldsymbol{\Theta}_s$ on both sides, and we have:
\begin{align}
\frac{\partial S}{\partial \boldsymbol{\Theta}_s}(\boldsymbol{x}_{\boldsymbol{\Theta}_s})+\left(\frac{\partial S}{\partial \boldsymbol{x}}\right)^T\frac{\partial \boldsymbol{x}}{\partial \boldsymbol{\Theta}_s }=0. ~\label{eq:implicit_diff}
\end{align}
Differentiating both sides of Eq.~\ref{eq:parametrize} with respect to $\boldsymbol{\Theta}_s$ gives:
\begin{align}
    \frac{\partial \boldsymbol{x}}{\partial \boldsymbol{\Theta}_s }={\boldsymbol{n}}\frac{\partial t}{\partial \boldsymbol{\Theta}_s }.\label{eq:parametrize_derivative}
\end{align}
Substituting Eq.~\ref{eq:parametrize_derivative} into Eq.~\ref{eq:implicit_diff} gives us:
\begin{align}
\frac{\partial S}{\partial \boldsymbol{\Theta}_s}(\boldsymbol{x}_{\boldsymbol{\Theta}_s})+\left(\frac{\partial S}{\partial \boldsymbol{x}}\right)^T{\boldsymbol{n}}\frac{\partial t}{\partial \boldsymbol{\Theta}_s }=0. ~\label{eq:implicit_diff3}
\end{align}
Evaluating Eq.~\ref{eq:implicit_diff3} at $\boldsymbol{\Theta}_s^{(0)}$, substituting $\boldsymbol{n}=\frac{\partial S}{\partial \boldsymbol{x}}\bigg\rvert_{\boldsymbol{\Theta}_s=\boldsymbol{\Theta}_s^{(0)}}$ and Eq.~\ref{eq:edge_point_value}, we have:
\begin{align}
\frac{\partial S}{\partial \boldsymbol{\Theta}_s}(\boldsymbol{x})\bigg\rvert_{\boldsymbol{\Theta}_s=\boldsymbol{\Theta}_s^{(0)}}+\boldsymbol{n}^T{\boldsymbol{n}}\frac{\partial t}{\partial \boldsymbol{\Theta}_s }\bigg\rvert_{\boldsymbol{\Theta}_s=\boldsymbol{\Theta}_s^{(0)}}=0. ~\label{eq:implicit_diff2}
\end{align}
Rearranging a bit, we have:
\begin{align}
    \frac{\partial t}{\partial \boldsymbol{\Theta}_s }\bigg\rvert_{\boldsymbol{\Theta}_s=\boldsymbol{\Theta}_s^{(0)}}=-\frac{1}{\boldsymbol{n}^T{\boldsymbol{n}}}\cdot \frac{\partial S}{\partial \boldsymbol{\Theta}_s}(\boldsymbol{x})\bigg\rvert_{\boldsymbol{\Theta}_s=\boldsymbol{\Theta}_s^{(0)}}.\label{eq:first_order}
\end{align}

Finally, we conclude our proof by observing that Eqns.~\ref{eq:zero_order},\ref{eq:first_order} gives us the first order approximation of $t_{\boldsymbol{\Theta}_s}$:
\begin{align}
    t_{\boldsymbol{\Theta}_s}=-\frac{1}{\boldsymbol{n}^T{\boldsymbol{n}}}\cdot S_{\boldsymbol{\Theta}_s}(\boldsymbol{x}).\label{eq:param_t}
\end{align}
Substituting Eq.~\ref{eq:param_t} into Eq.~\ref{eq:parametrize} leads to our proposed edge point re-parametrization:
\begin{align}
        \boldsymbol{x}_{\boldsymbol{\Theta}_s}=\boldsymbol{x}-\frac{\boldsymbol{n}}{\boldsymbol{n}^T{\boldsymbol{n}}}\cdot S_{\boldsymbol{\Theta}_s}(\boldsymbol{x}).
\end{align}
We note that $\boldsymbol{x}_{\boldsymbol{\Theta}_s}$ only provides unbiased first-order gradient with respect to $\boldsymbol{\Theta}_s$ suitable for gradient-based optimizers.

\section{Reconstructed meshes and materials}
In Fig.~\ref{fig:assets}, we show the reconstructed meshes and materials for the 5 real-world scenes used in this work.

\section{Comparison with PhySG}
\begin{figure}[htp]
    \centering
    \includegraphics[width=0.95\columnwidth]{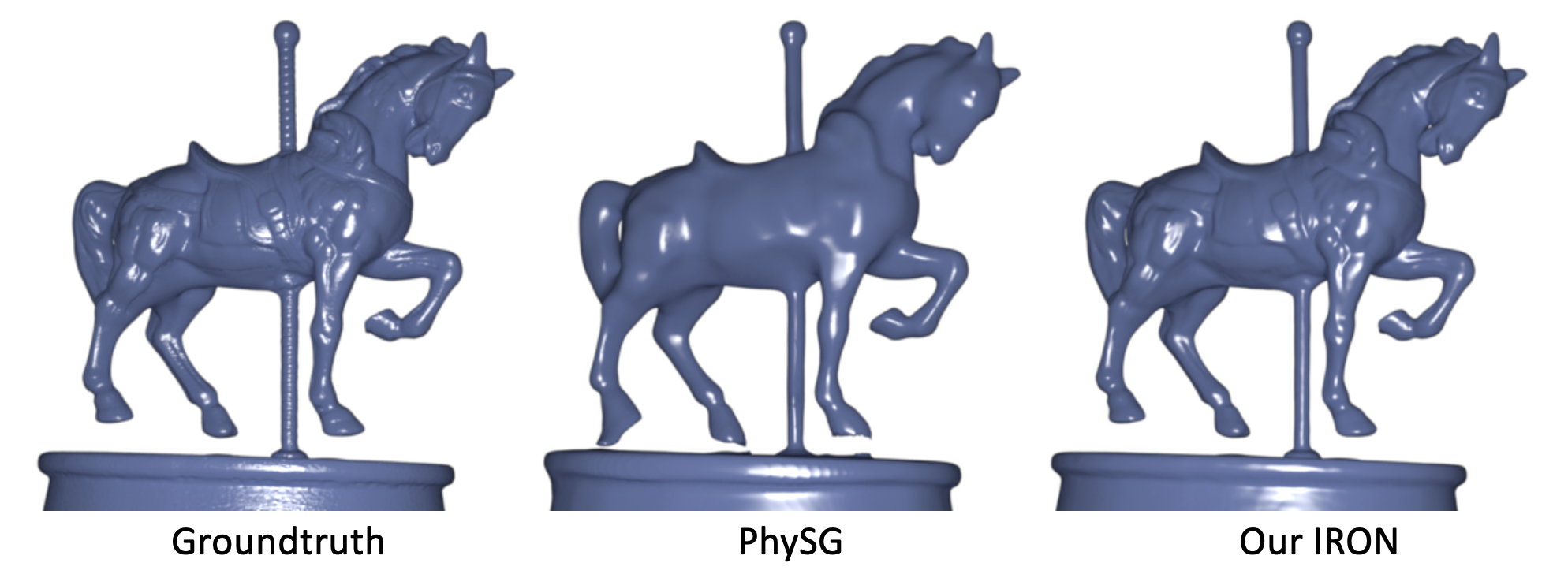}
    \caption{For the horse object, our method recovers much more accurate geometry details than PhySG; chamfer L1 distances are:  Our IRON (5.35e-4) vs. PhySG (18.67e-4).}
    \label{fig:compare_with_physg}%
\end{figure}

First, note there are a few key limitations to PhySG that our method can overcome. Namely, unlike our method, PhySG requires input object segmentation masks. Our method can also handle spatially-varying specular roughness, whereas PhySG assumes a constant and uniform specular lobe shape.
Because PhySG assumes static environmental lighting, we used Mitsuba to render the same object from the same set of viewpoints first using a collocated flash (to run our method), and then with environmental lighting (to run PhySG).  See Fig.~\ref{fig:compare_with_physg} and the caption for an example comparison. 

{\small
\bibliographystyle{ieee_fullname}
\bibliography{refs}
}

\end{document}